\documentclass[11pt]{article}

\usepackage[letterpaper,margin=1in]{geometry}
\usepackage[T1]{fontenc}
\usepackage{times}
\usepackage{microtype}
\usepackage{graphicx}
\usepackage{subcaption}
\usepackage{booktabs}
\usepackage{multirow}
\usepackage{xcolor}
\usepackage[round]{natbib}
\usepackage{amsmath}
\usepackage{amssymb}
\usepackage{mathtools}
\usepackage[colorlinks=true,linkcolor=blue!55!black,citecolor=blue!55!black,urlcolor=blue!55!black]{hyperref}
\usepackage{xurl}
\usepackage[capitalize,noabbrev]{cleveref}
\usepackage{siunitx}
\usepackage{caption}
\title{SketchMate: Agentic Geometric Reasoning for\\Zero-Shot Precision Insertion}
\author{Jiaming Wang}
\date{}

\begin{document}

\maketitle
\begin{center}
  \includegraphics[width=\textwidth]{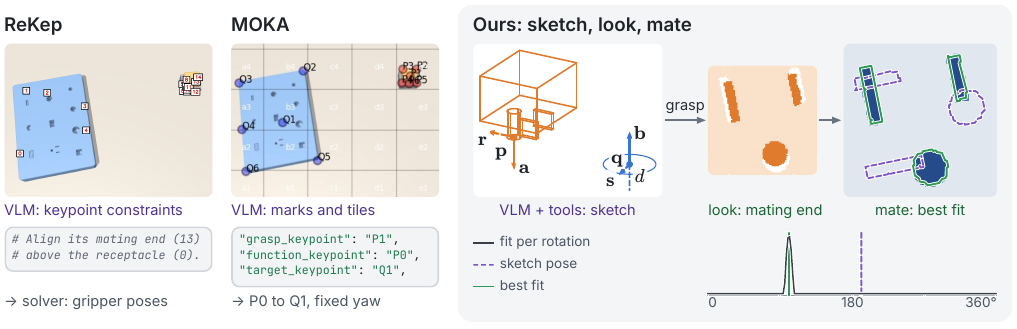}
  \captionof{figure}{\textbf{VLMs know which part goes where, and geometry decides how it mates.} Real outputs of three methods for inserting a NEMA 5-15 plug into a task board, with the source of each motion in grey. ReKep~\citep{rekep} lets the VLM write constraints between image keypoints, and MOKA~\citep{moka} lets it select marks and tiles. Our agent builds a metric sketch with geometric tools (orange: moving part, blue: fixed part, symbols as in \cref{sec:sketch}). This sketch identifies the correct opening but misjudges the rotation of the plug by \SI{90}{\degree} (purple, dashed). Once the plug is grasped, matching close-up views of both mating faces recovers the correct rotation (green).}
  \label{fig:teaser}
\end{center}

\begin{abstract}
Inserting an unseen part requires sub-millimetre precision and the correct orientation of keyed features. Current zero-shot agentic manipulation methods built on vision-language models (VLMs) produce spatial outputs that are far too coarse for this. We propose to let a VLM build a \emph{sketch model} of the task, i.e., a few geometric primitives and two mate frames, which a VLM agent measures with geometric tools in multi-view depth and which a standard planner and compliant controller execute. Because a sketch describes the parts independently of their placement, the robot needs to build it only once per task. A sketch memory stores each agent run and transfers the stored sketch to later placements by point-cloud registration, so that the VLM is called only for unfamiliar parts. After grasping the part, the robot inspects both mating faces at close range and corrects the sketch by shape matching. On 69 simulated insertion episodes, our method inserts 88\% of the parts at the assembled pose in a strict kinematic test, while ReKep and MOKA with the same VLM insert none. On new placements of eight tasks, reusing one stored run inserts as many parts as calling the agent in each episode, and it produces the sketch in \SI{3.4}{s} instead of about \SI{90}{s}.
\end{abstract}

\section{Introduction}
\begin{figure}[t]
  \centering
  \includegraphics[width=\textwidth]{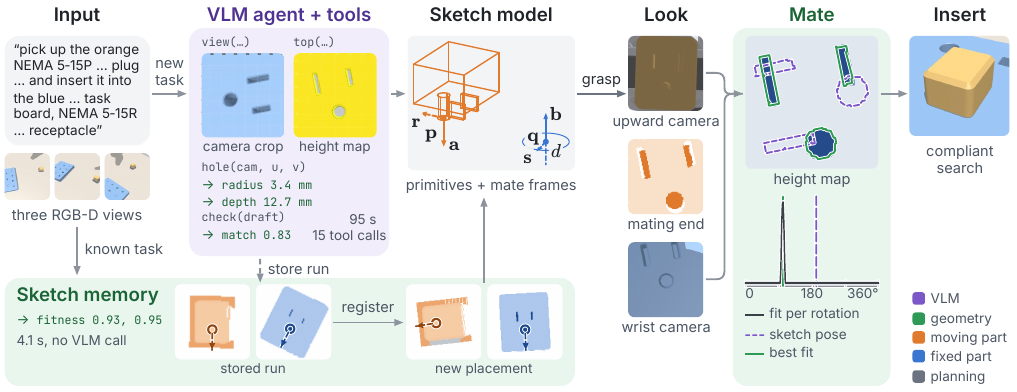}
  \caption{\textbf{Overview.} A VLM agent builds the sketch for a new task, and the sketch memory supplies it for a known task. The agent (purple) calls geometric tools on three RGB-D views, whereas the memory registers a stored run to the new scene by geometry alone (green). After the grasp, we match the mating end (dark orange) to the height map of the fixed part and insert the part with compliance. The panels show one episode with GPT-6.1-Sol and the NEMA 5-15 task of \cref{sec:memoryexp}, with symbols as in \cref{sec:sketch}.}
  \label{fig:overview}
\end{figure}

Insertion is at the heart of robotic assembly, whether a plug goes into a receptacle, a transceiver into a network card or a peg into a fixture. It is also the step that resists generalization. Typical parts mate with clearances of \SIrange{0.1}{0.3}{mm} and tolerate rotational errors of only one or two degrees. Many of them are also keyed, which means that they fit in a single orientation. For this reason, industrial practice engineers each new part individually with CAD models, fixtures and manually taught poses, and learning-based methods train a dedicated policy for each assembly or family of assemblies~\citep{industreal,automate,srsa}. In this paper, we investigate whether a robot can instead insert an unseen part, given only camera observations and a natural-language instruction such as ``insert the USB plug into the task board''.

Vision-language models (VLMs) are a natural source of such generality, since they know what a USB plug is and which receptacle it belongs to. In agentic robotics, VLMs and other foundation models make at run time some of the decisions that engineers otherwise make before deployment~\citep{agenticsurvey}. However, current zero-shot manipulation methods ground this knowledge in coarse spatial representations, such as marks and image tiles~\citep{moka}, keypoints on image patches with constraints written by the VLM~\citep{rekep}, or value maps on a voxel grid~\citep{voxposer}. The resolution of these representations is limited by patches and voxels of several millimetres. In addition, these methods ask the VLM to infer 3D correspondences from an image, for example which point of the plug has to meet which point of the receptacle.

We find that these representations are insufficient for insertion. When we port ReKep and MOKA to our tasks with our VLM, planner and controller, they insert none of the 69 parts at the assembled pose (\cref{sec:main}).

Our approach assigns the semantic decisions to the VLM and the measurements to geometry. We let the VLM construct a \emph{sketch model} of the task, which consists of a few geometric primitives covering the moving part and two mate frames. These frames locate the mating end of the moving part and specify which opening of the fixed part it enters, or which post it slides onto, and in which orientation (\cref{fig:teaser}). The VLM builds the sketch iteratively in an agent loop, using geometric tools that operate on multi-view depth data. In this division of labour, the VLM decides what to measure, and the tools provide the precision. For example, the VLM points at the rim of a hole, and a tool returns the circle fitted to the corresponding 3D points. A standard motion planner and a compliant controller then execute the sketch.

Running the agent takes roughly \SI{90}{s} and tens of thousands of tokens per episode, which is costly to repeat whenever the same parts reappear. However, a sketch describes the parts themselves independently of their placement, which means that the robot needs to construct it only once per task. We store each agent run in a sketch memory along with the point clouds of both parts. In a later episode, we register these point clouds to the new scene and transfer the stored sketch, so that the VLM is consulted only for parts that the memory fails to recognize.

Keyed parts reveal a second problem. The agent reliably identifies which part goes into which opening, but it frequently misreads the rotation of the moving part, for instance the side of a USB plug that carries the tongue (\cref{sec:analysis}). More precise measurements cannot correct this correspondence error, because the VLM itself chooses the measurements.

However, the correspondence becomes a geometric problem as soon as both mating faces are visible. After grasping the part, our robot therefore holds its mating end over an upward-facing camera, views the fixed part at close range with a wrist camera, and searches for the pose in which the two shapes best complement each other. Because both views are close, the same search also corrects offsets below the resolution of the scene cameras, as well as the registration errors of reused sketches. The sketch still provides the opening, the axis and the depth, and it resolves ties between fits that are geometrically equivalent.

We evaluate our method in simulation on assemblies from AutoMate~\citep{automate}, connectors built to their published dimensions, a NIST-style task board~\citep{nistboard}, the parts of the AI for Industry Challenge~\citep{aic} and six held-out connector types. Our method inserts 61 of the 69 development parts at the assembled pose in a strict kinematic test and 91\% in physics simulation. In contrast, ReKep and MOKA insert none, because their keypoints and marks do not describe the mating features. The look-and-mate step corrects the rotations that the agent misreads as well as offsets below the resolution of the scene cameras, and on the held-out connectors, the complete method inserts all 18 parts in both tests. On new placements of eight tasks, reusing the sketch of a single agent run inserts 30 of 32 parts with registration alone, which matches the performance of calling the agent in each episode.

This paper makes three contributions. First, we introduce sketch models, a task representation that a VLM agent builds once per task with geometric tools, that a sketch memory transfers to new placements by registration and that a classical planner executes. Second, we propose a look-and-mate step that observes both mating faces at close range after the grasp and resolves their correspondence by shape complementarity under the semantic constraints of the sketch. Third, we evaluate our method against ported VLM baselines and report, besides success, the accuracy of each intermediate representation and the cause of each failure.

\section{Related Work}

\paragraph{VLMs for robot manipulation.}
Vision-language-action models map images and instructions directly to actions~\citep{rt2,openvla,pi0}, and to our knowledge none has been shown to reach sub-millimetre precision on unseen parts. Zero-shot methods instead let a pretrained VLM produce an intermediate spatial representation that a solver converts into motion. Examples include programs over perception APIs~\citep{codeaspolicies}, value maps~\citep{voxposer}, points or marks in the image~\citep{moka,pivot,robopoint}, part constraints~\citep{copa}, keypoint constraints~\citep{rekep} and interaction frames~\citep{omnimanip}. These representations are anchored on image features of several millimetres, and they require the VLM to infer 3D relations, a known weakness of current models~\citep{whatsup,blink,mmvp,spatialvlm}. In our agent, the VLM chooses the measurements, and geometry computes the relation that VLMs estimate least reliably, namely the rotational correspondence of keyed parts.

\paragraph{Robotic insertion and assembly.}
Classical insertion assumes known part models and poses, and it absorbs small errors with compliance and search~\citep{whitney,chhatpar}. Learning-based methods train insertion policies in simulation or on the real robot for each assembly or family of assemblies~\citep{factory,industreal,automate,srsa,serl}. These methods achieve high success rates, but they require training as well as the part geometry or poses. Similarly, pipelines based on 6D pose estimation~\citep{foundationpose,megapose} need a CAD model and the mating transformation of each part. Benchmarks such as the NIST task boards~\citep{nistboard} and FMB~\citep{fmb} show how many connector types a general system has to handle. In contrast, we target an unseen pair of parts that is specified by an instruction alone and observed with RGB-D cameras, and we retain the compliant search of classical insertion as the final step.

\section{Method}
\label{sec:method}

In this section, we first formalize the problem (\cref{sec:problem}) and define sketch models (\cref{sec:sketch}). A sketch originates either from a VLM agent for a new task (\cref{sec:agent}) or from a sketch memory for a known task (\cref{sec:memory}). As shown in \cref{fig:overview}, both sources feed the same execution pipeline. In both cases, the robot grasps the moving part (\cref{sec:plan}), observes both mating faces at close range, corrects the sketch by shape matching and finally inserts the part (\cref{sec:look}).

\subsection{Problem Statement}
\label{sec:problem}
A robot has to pick up a \emph{moving part} $A$ and mate it with a \emph{fixed part} $B$, for example a plug with its receptacle, a peg with its hole or a ring with a post. Initially, $A$ stands on the table or in a simple holder, and $B$ is mounted on the table. Box-shaped obstacles may stand between them. The robot observes the scene with three calibrated RGB-D cameras, carries a wrist camera and can present a part to an upward-facing camera embedded in the table. A natural-language instruction identifies the two parts, for example ``pick up the orange USB Type-A plug and insert it into the blue task board, USB Type-A receptacle''. The robot has to accomplish the task from these observations and the instruction alone, without CAD models, demonstrations or task-specific training.

Let $X\in SE(3)$ denote the pose of $A$ relative to $B$, and let $\mathcal{X}^\star$ denote the set of assembled poses, in which the mating features are fully engaged. $\mathcal{X}^\star$ contains $n$ discrete poses for a mate with $n$-fold rotational symmetry and a continuum for a round mate. The robot has to grasp $A$ and transport it along a collision-free path to within the tolerance of some $X^\star\in\mathcal{X}^\star$. For our procedural parts, this tolerance is \SIrange{0.2}{0.45}{mm} laterally and \SIrange{1.3}{4.6}{\degree} in rotation (\cref{app:tasks}). We consider mates that are reached by a straight motion along a single axis, either into an opening or onto a post. Tasks recur, since the same two parts can reappear in many episodes with different placements and obstacles.

\begin{figure}[t]
  \centering
  \includegraphics[width=\textwidth]{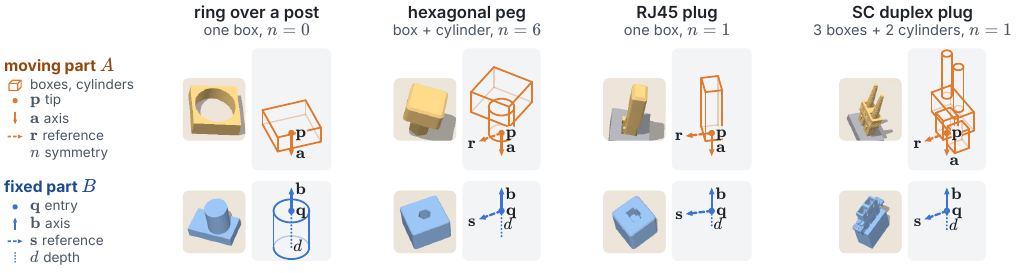}
  \caption{\textbf{A sketch model is a few primitives and two mate frames.} Agent sketches for four tasks beside crops of the camera images that the agent received (orange: moving part $A$, blue: fixed part $B$). Primitives of $B$ are optional, and the agent added one only for the post. Each $B$ is turned so that $\mathbf{s}\parallel\mathbf{r}$, as in the assembled state, and each sketch alone inserts the part.}
  \label{fig:sketches}
\end{figure}

\subsection{Sketch Models}
\label{sec:sketch}
A sketch model is the minimal geometric description of a task that a motion planner requires to grasp $A$ and insert it. We denote it by $\mathcal{S}=(\mathcal{S}_A,\mathcal{S}_B)$, expressed in the world frame. The moving-part sketch $\mathcal{S}_A=(\Pi,\mathbf{p},\mathbf{a},\mathbf{r},n)$ contains a set $\Pi$ of primitives that jointly cover $A$, as well as a mate frame. This frame consists of the tip $\mathbf{p}$ at the centre of the mating end, the insertion axis $\mathbf{a}$ along which $A$ approaches the mate, and a reference direction $\mathbf{r}\perp\mathbf{a}$ that determines the rotation of $A$ around this axis. The integer $n$ denotes the order of rotational symmetry of the mate, with $n=0$ for a round mate. Each primitive is either an upright box parameterized by centre, extents and yaw angle or a cylinder parameterized by the centres of the two end faces and the radius. The fixed-part sketch $\mathcal{S}_B=(\mathbf{q},\mathbf{b},\mathbf{s},d)$ contains the entry $\mathbf{q}$ (the centre of the opening or the top of a post), the axis $\mathbf{b}$ pointing out of $B$, a reference direction $\mathbf{s}\perp\mathbf{b}$ and the insertion depth $d$ that the tip travels beyond the entry. Optionally, the agent also suggests a grip. \Cref{fig:sketches} shows sketches that the agent constructed for parts with different mates.

A sketch can be considerably coarser than the parts. The primitives are used only for grasp sampling and for collision checking of the carried part, and for this purpose a few boxes and cylinders suffice. A ring, for example, can be represented by one solid box. Primitives of the fixed part are optional, since the planner checks collisions against the observed point cloud. The assembled pose of $A$ is the rigid motion $(\mathbf{R},\mathbf{t})$ of the sketch that satisfies
\begin{equation}
\mathbf{R}\mathbf{a}=-\mathbf{b},\quad \mathbf{R}\mathbf{r}=\mathrm{Rot}\big(\mathbf{b},\tfrac{2\pi k}{n}\big)\,\mathbf{s},\quad \mathbf{R}\mathbf{p}+\mathbf{t}=\mathbf{q}-d\,\mathbf{b}
\label{eq:mate}
\end{equation}
for some integer $k$, i.e., the axes are opposed, the references agree up to the symmetry and the tip lies at depth $d$. For a round mate ($n=0$), the rotation around $\mathbf{b}$ is unconstrained.

\subsection{Building a Sketch with a VLM Agent}
\label{sec:agent}
For a new task, a VLM builds the sketch in a tool loop, as in tool-using language agents~\citep{react,visprog,vipergpt}. The VLM receives the instruction, overview images with labelled objects and a list of the segmented objects. We implement the tools on a geometric backend that fuses the depth images into one point cloud, removes the robot and the table, and splits the remaining points into objects by Euclidean clustering, as in tabletop segmentation for mobile manipulation~\citep{rusu2009}. The system uses no learned segmentation model, because the VLM identifies the parts among these objects. Each tool answers one geometric question on these data (\cref{tab:tools}). For example, \texttt{hole} returns the rim of the opening around a pixel, and \texttt{top} returns an orthographic height map with a millimetre grid in which keys and slots appear free of perspective distortion. The tool \texttt{check} compares a draft sketch with the depth data. Our system prompt instructs the VLM to take numbers from the tools instead of the images and to verify the draft before answering. The VLM thus decides which part moves, which opening is meant, which features correspond and what to measure, while the tools return metric values (\cref{fig:tools}).

\begin{table}[t]
\caption{Tools of the agent. Each tool answers one geometric question on the fused depth data of the episode and returns JSON in millimetres in the world frame. \texttt{view}, \texttt{top} and \texttt{check} also return an image. Runs: share of the 141 agent runs that call the tool, with the median number of calls in these runs.}
\label{tab:tools}
\centering
\small
\begin{tabular}{l p{3.5cm} p{8.3cm} r}
\toprule
Tool & Input & Returns & Runs \\
\midrule
\texttt{view} & camera, optional pixel box & the image or a crop of it, with a grid labelled in camera pixels & 97\%, 4 \\
\texttt{xyz} & camera, pixels & the 3D point, normal and object at each pixel, and the near and far points at depth edges & 77\%, 2 \\
\texttt{region} & pixel or object & the object fused over all cameras: height range, upright box, vertical cylinder, heights of flat upward faces and radius profile & 100\%, 2 \\
\texttt{hole} & pixel inside an opening & the rim of the opening: centre, circle radius, rectangle, rim height and the depth visible below the rim & 100\%, 2 \\
\texttt{top} & square area, height range & an orthographic height map with a millimetre grid, in which keys, flats and slots appear without perspective & 98\%, 3 \\
\texttt{fit} & 3D points, model type & a plane, circle, line, cylinder, upright box or rigid pose fitted to the points & 24\%, 2 \\
\texttt{check} & draft sketch, optional drawing & schema errors, the agreement of each primitive with the depth data, the share of the observed part that the primitives explain, and optionally the sketch drawn on an image & 100\%, 1 \\
\bottomrule
\end{tabular}
\end{table}

\subsection{Reusing Sketches without the VLM}
\label{sec:memory}
A sketch describes the two parts in their own geometry and remains valid when they are placed elsewhere. We store each agent run in a sketch memory and reuse it on subsequent placements of the same task. For each run, the memory stores the observed points of $A$ and $B$ in object frames on the table (the object templates), the sketch expressed in these frames and the colour of each part. In a new episode, we segment the scene identically and register the templates of each stored run to the objects of similar colour and size. Since the parts rest on the table, this registration is a planar rigid transformation in $x$, $y$ and yaw. We estimate this transformation using a coarse yaw sweep followed by ICP~\citep{icp}. Its fitness is the fraction of object points close to the template (\cref{app:impl}). We transform the sketch of the best-fitting run into the scene, $\mathcal{S}'=(\mathbf{T}_A\mathcal{S}_A,\mathbf{T}_B\mathcal{S}_B)$, where $\mathbf{T}_A$ and $\mathbf{T}_B$ denote the registrations of the two parts. If the fitness of either part falls below a threshold, we call the agent instead. Like one-shot imitation methods that retrieve and align the closest demonstration~\citep{dinobot}, the memory requires no training, but it stores sketches built by the agent instead of human demonstrations. We use a single run instead of an average over several runs, because the frames of different runs can disagree, for example when a nearly round part registers ambiguously in yaw (\cref{sec:memoryexp}).

\subsection{Grasping, Planning and Insertion}
\label{sec:plan}
Given a sketch from the agent or the memory, a standard planner samples top-down grasps across the primitives and plans collision-free paths with RRT-Connect~\citep{rrtconnect}. The planner prefers the grip that the agent suggests. The placement pose follows from \cref{eq:mate} and the in-hand pose of $A$. Because the look (\cref{sec:look}) measures $A$ in the gripper, we plan the insertion path after the look. During grasp selection, the planner therefore checks only that the sketched mate pose is reachable and collision-free. A task-space impedance controller then lowers the part along the axis. When the part is blocked at the entry, the controller performs a spiral and a rotary search as in classical peg-in-hole insertion~\citep{chhatpar} (\cref{app:impl}). Small offsets can be absorbed by compliance and search~\citep{whitney}, but not an incorrect rotation or opening.

\subsection{Look, Then Mate}
\label{sec:look}
The sketch can be wrong in ways that compliance cannot absorb. The agent frequently misreads the rotation of keyed parts (\cref{sec:analysis}). The scene cameras resolve openings only to about \SI{0.4}{mm}, which exceeds typical clearances, and a reused sketch adds the registration error. When both mating faces are viewed at close range, however, the correspondence becomes a well-posed geometric problem that we solve for all parts by geometry alone (\cref{sec:ablation}).

\begin{figure}[t]
  \centering
  \includegraphics{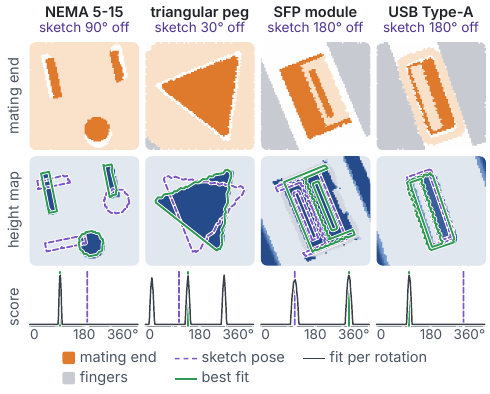}
  \caption{\textbf{Geometry recovers the rotations that the VLM misreads.} Each column shows the mating end seen by the upward camera (dark orange, fingers in grey), the height map of the fixed part with this end at the sketch pose (purple, dashed) and at the best fit (green), and the best score per rotation. A symmetric mate gives equal peaks (triangle), and a flip that differs only in a small key scores lower (SFP).}
  \label{fig:fixes}
\end{figure}

\paragraph{Profile of the moving part.} After grasping $A$, the robot holds the mating end above the upward-facing camera, as component placement machines in electronics manufacturing do. From this view, we compute the profile of $A$ in the gripper frame as its lower envelope. For each cell of a fine grid perpendicular to the insertion axis, the profile holds the point of $A$ that lies furthest along the axis. The profile also includes the fingers, which move with $A$ (\cref{app:impl}). In addition, this view measures the actual in-hand pose of $A$.

\paragraph{Height map of the fixed part.} The robot then positions the wrist camera above the sketched entry, looking along $\mathbf{b}$ from a distance at which one pixel covers \SI{0.16}{mm}. In a frame centred at $\mathbf{q}$ with $z$ along $\mathbf{b}$, we record for each grid cell the height $h(x,y)$ of the highest observed surface, which yields the upper envelope of $B$. The bottom of a narrow opening is frequently occluded in all cameras. We therefore mark a cell inside the outline of $B$ as open when camera rays traverse it, similar to free-space carving in volumetric range-image fusion~\citep{curless}. At height $z$, the solid region of $B$ is then $\{(x,y): h(x,y)>z\}$.

\paragraph{Mating by shape complementarity.} A candidate mate consists of a rotation $\psi$ of $A$ around $\mathbf{b}$ and a position $\mathbf{t}$ of its tip in the entry plane. We score each candidate by how well the profile complements the height map when lowered to several insertion depths $\ell\in\mathcal{L}$,
\begin{equation}
\begin{aligned}
S(\psi,\mathbf{t})=\sum_{\ell\in\mathcal{L}}\Big[&\sum_{c\in\partial P}\mathbb{1}\big[\omega_{z_c}(\mathbf{x}_c)\le\beta\big]\\[-2pt]
&-\sum_{c\in P}\big[\delta_{z_c}(\mathbf{x}_c)-\tau\big]_+/\Delta\Big],
\end{aligned}
\label{eq:score}
\end{equation}
where $P$ denotes the set of profile cells, $\partial P$ its outline, and $\mathbf{x}_c$ and $z_c$ the position and height of cell $c$ for the given candidate and depth. The penetration $\delta_z(\mathbf{x})$ measures the distance from a solid point at height $z$ to the nearest free point, i.e., how deeply a profile cell penetrates $B$, and the clearance $\omega_z(\mathbf{x})$ measures the distance from a free point to the nearest wall. $\Delta$ denotes the grid resolution, $\tau$ a penetration tolerance, and $\beta$ the largest gap at which an outline cell still counts as touching a wall (\cref{app:impl}). The first term rewards outline cells that touch a wall while remaining in free space, and the second term penalizes material of $A$ that penetrates $B$. As in protein docking~\citep{katchalski}, a part inside its opening is surrounded by walls, unlike a part beside the opening or resting on top of it. Because the score counts cells instead of taking a maximum, it tolerates fits tighter than the grid resolution as well as isolated depth errors.

We search exhaustively over rotations and, for each rotation, over positions in a small window around the sketched pose. This window bounds the centre of the outline of $A$ instead of its tip, because a misread rotation turns the sketch about the centre of the part (\cref{app:impl}). Geometry then eliminates the candidates that penetrate $B$ and ranks the remaining ones by contact. Several fits can come close to the best one, for example symmetric rotations, identical ports or a part that can slide along a bar. Among these equivalent fits, we select the one closest to the rotation and position of the sketch. If the sketched rotation is far from all fits, we treat it as a misreading and select the best-scoring fit. The sketch thus supplies the opening, the axis and depth, the search window and the choice among equivalent fits, while geometry supplies the rotation and the sub-millimetre position. The opening has to come from the sketch, because shape matching alone selects a looser receptacle in 3 of the 10 board tasks (\cref{app:impl}). The robot finally inserts $A$ at the matched pose, along a path planned from the close observation.

\section{Experiments}
\label{sec:exp}

We first compare all methods and report the main results (\cref{sec:main}). We then analyze their failures and intermediate representations (\cref{sec:analysis}), study the contribution of the look and its close view (\cref{sec:ablation}) and evaluate the reuse of sketches (\cref{sec:memoryexp}).

\subsection{Setup}
\label{sec:setup}

\begin{table}[t]
\caption{Insertion success (number of episodes) in the kinematic test, strict or within reach of the compliant search, and in physics from valid starts (\cref{sec:setup}). All methods share GPT-6.1-Sol, the planner and the controller, and Oracle runs our method on ground-truth sketches. $^\dagger$Search-range episodes and ten random others. $^\ddagger$Valid in this run and ours.}
\label{tab:main}
\centering
\small
\setlength{\tabcolsep}{4pt}
\begin{tabular}{l ccccc c c cc}
\toprule
 & \multicolumn{6}{c}{Kinematic, strict (development)} & Held-out & \multicolumn{2}{c}{Physics} \\
\cmidrule(lr){2-7}\cmidrule(lr){8-8}\cmidrule(lr){9-10}
Method & AutoMate & Conn. & Board & AIC & All & Search & strict & Devel. & Held-out \\
 & (42) & (13) & (10) & (4) & (69) & range & (18) & (66) & (18) \\
\midrule
ReKep~\citep{rekep} & 0 & 0 & 0 & 0 & 0 & 0 & 0 & 0/10$^\dagger$ & -- \\
MOKA~\citep{moka} & 0 & 0 & 0 & 0 & 0 & 2 & 0 & 0/12$^\dagger$ & -- \\
\midrule
Ours, sketch only & 32 & 11 & 6 & 1 & 50 & 60 & 6 & 54 & 10 \\
Ours & \textbf{35} & \textbf{13} & \textbf{9} & \textbf{4} & \textbf{61} & \textbf{64} & \textbf{18} & \textbf{60} & \textbf{18} \\
\midrule
\textcolor{gray}{Oracle} & \textcolor{gray}{37} & \textcolor{gray}{13} & \textcolor{gray}{10} & \textcolor{gray}{4} & \textcolor{gray}{64} & \textcolor{gray}{68} & \textcolor{gray}{18} & \textcolor{gray}{61/65$^\ddagger$} & \textcolor{gray}{18} \\
\bottomrule
\end{tabular}
\end{table}

\paragraph{Tasks.} All our tasks are pick-and-insert episodes in simulation with a Franka arm, with both parts at random positions and yaws and up to two box obstacles (\cref{app:tasks}). \textbf{AutoMate} contains the 42 assemblies of AutoMate~\citep{automate} where the moving part stands stably on its mating end. \textbf{Connectors} contains 13 procedural parts built to their published dimensions, namely power plugs, USB Type-A, RJ45, D-sub 9 and keyed or symmetric pegs, including the rectangular peg of Factory~\citep{factory}. \textbf{Board} is a task board in the style of NIST ATB1~\citep{nistboard} with ten receptacles. \textbf{AIC} contains two placements each of the SFP transceiver and the SC fiber connector of the AI for Industry Challenge~\citep{aic}. \textbf{Held-out} contains three placements each of six further connector types, namely USB Type-C, HDMI, DisplayPort, XLR, the Europlug and a DC barrel jack, which we held out from development. The procedural parts have clearances of \SIrange{0.12}{0.3}{mm} per side (\cref{app:tasks}).

\paragraph{Perception.} Three stationary RGB-D cameras observe the workspace from a distance of \SIrange{0.6}{0.7}{m} at a resolution of $1280\times960$ pixels. At the fixed part, the camera with the finest resolution covers a median of \SI{0.38}{mm} per pixel. The upward-facing and wrist cameras have identical resolution. To isolate the effect of the task representation, we render noise-free depth (\cref{sec:limits}).

\paragraph{Metrics.} We evaluate each method with two complementary tests. The \emph{kinematic} test executes the plan with a rigid grasp, including the insertion path planned after the look. It counts the moving part as inserted if, along the entire insertion path, its penetration into the fixed part is at most that of the assembled CAD models and the fixed part constrains it laterally (a sideways displacement of \SI{1.5}{mm} collides). We report \emph{strict} success, within \SI{2}{mm} and \SI{5}{\degree} of the assembled pose, and whether the final pose lies within reach of the compliant search (\SI{3}{mm}, \SI{10}{\degree}). This test is deterministic but conservative, since compliance can absorb errors of a few tenths of a millimetre.

The \emph{physics} test simulates the complete episode in Isaac Lab~\citep{isaaclab} with the contact dynamics of AutoMate. It counts a part as inserted when its final position lies within \SI{2}{mm} of the fixed-part axis and at most \SI{1.5}{mm} above its assembled height, with an orientation within \SI{5}{\degree} of upright and of a valid rotation. We exclude episodes where the moving part topples before the robot moves.

\paragraph{Baselines.} We compare with two published zero-shot VLM manipulation methods with released code, which we port to our episodes. ReKep~\citep{rekep} proposes keypoints from DINOv2 features~\citep{dinov2} inside object masks, asks the VLM to write a program of keypoint constraints and solves for end-effector subgoals by constrained optimization. MOKA~\citep{moka} asks the VLM to select grasp, function and target points and waypoints among marks drawn on the image. We keep their prompts, keypoint proposals, parsers and solvers and give them the object masks of the simulator. Our planner and compliant controller execute their poses, with a guarded push beyond their target, since neither method predicts an insertion depth (\cref{app:baselines}). For all methods, we use the same VLM, GPT-6.1-Sol with medium reasoning effort. As a reference, \emph{Oracle} runs our method on ground-truth sketches computed from the CAD models.

\subsection{Main Results}
\label{sec:main}

As shown in \cref{tab:main}, our method inserts 61 of the 69 development parts at the assembled pose in the strict test and 60 of 66 in physics. In contrast, ReKep and MOKA insert none in either test. The two baselines fail long before precision matters, because their final poses lie a median of \SI{25}{mm} (ReKep) and \SI{13}{mm} (MOKA) from the opening and are often rotated (\cref{tab:analysis}). Since all methods share the same planner and controller, these failures likely originate from their representations. ReKep proposes keypoints from image features, which on the task board of \cref{fig:teaser} are located on the outline and corners of the board, away from the NEMA receptacle. MOKA moves a mark on the moving part to a mark on the outline or centroid of the fixed part at a fixed yaw, and thus cannot represent the rotation of a keyed part.

We find that the sketch models carry the semantics and most of the precision. The sketch alone inserts 50 of the 69 parts at the assembled pose and brings 60 within reach of the compliant search, with a median final error of \SI{0.15}{mm}. The look adds eleven parts without losing any, and it lowers the median error to \SI{0.10}{mm}. These additional parts are keyed parts with misread rotations and parts with openings that the scene cameras resolve too coarsely (\cref{sec:ablation}). With ground-truth sketches, the same pipeline inserts 64 parts, which means that the sketches built by the agent lose only the four parts where it selected an incorrect feature or receptacle. In physics, the compliant search absorbs most of the remaining sub-millimetre errors, and the sketch alone inserts 54 of the 66 parts.

The results on the held-out connectors follow the same trends. The complete method inserts all 18 parts in both tests, compared with 6 and 10 for the sketch alone, respectively. The look-and-mate step corrects all seven rotations that the agent misread, including those of the three XLR plugs, as well as five offsets of \SIrange{0.2}{0.5}{mm}.

\subsection{Where Do the Methods Fail?}
\label{sec:analysis}

\begin{table}[t]
\caption{Failure causes (development episodes). Median lateral error of the final pose (mm) and the cause of each strict failure: more than \SI{3}{mm} from the opening (place), more than \SI{5}{\degree} from any fitting rotation (rotation), a smaller miss (imprecise) or the wrong object grasped (grasp).}
\label{tab:analysis}
\centering
\small
\setlength{\tabcolsep}{2.5pt}
\begin{tabular}{l c ccccc}
\toprule
 & Median & \multicolumn{5}{c}{Episodes (of 69)} \\
\cmidrule(lr){3-7}
Method & error & success & place & rotat. & impr. & grasp \\
\midrule
ReKep & 25 & 0 & 54 & 1 & 0 & 14 \\
MOKA & 13 & 0 & 67 & 1 & 1 & 0 \\
Sketch only & 0.15 & 50 & 3 & 6 & 10 & 0 \\
Ours & 0.10 & 61 & 4 & 1 & 3 & 0 \\
\textcolor{gray}{Oracle} & \textcolor{gray}{0.01} & \textcolor{gray}{64} & \textcolor{gray}{0} & \textcolor{gray}{1} & \textcolor{gray}{4} & \textcolor{gray}{0} \\
\bottomrule
\end{tabular}
\end{table}

\Cref{tab:analysis} attributes each strict failure to a single cause. ReKep frequently grasps an incorrect object, because the VLM mistakes a keypoint on the holder or on the fixed part for a keypoint on the moving part (\cref{app:failures}).

Our sketches fail for three reasons (\cref{app:failures}). First, six episodes have an incorrect rotation, all of them on keyed parts. For example, the VLM rotated the NEMA 5-15 plug on the board by \SI{90}{\degree}. The look recovers four of these rotations (\cref{fig:fixes}). It misses two AutoMate parts. One is keyed by an internal post, which hides its walls from the score, and for the other, the agent selected the incorrect one of two barrels. Second, ten episodes are imprecise by \SIrange{0.1}{1}{mm}. The close look corrects seven of them, and the remaining three fail the strict test even with ground-truth sketches. Third, three episodes place the moving part at an incorrect location, such as the USB-A plug, sent by the agent to a larger receptacle that accepts it loosely.

To quantify how accurately the agent estimates rotations, we pool all 86 episodes with a keyed part and an agent sketch, including the held-out and reuse placements. The sketched rotation falls within the task tolerance in 57 of these episodes and the rotation after the look in 83. This correction raises strict success from 36 to 82 episodes (46 gained and 0 lost, $p<10^{-13}$, exact McNemar test~\citep{mcnemar}). The sketch errors concentrate on parts that the agent misreads in most placements, such as the NEMA 5-15 plug in 7 of 8 placements and the XLR plug in all three. This pattern suggests that the VLM holds a consistent but incorrect belief about these parts.

The look-and-mate step itself fails in four of these episodes. In two SC placements, the sketch is flipped, and both rotations of this duplex connector fit the opening, because they differ only in a key below the camera resolution. The remaining two failures are the USB-A plug in an incorrect receptacle and an RJ45 plug that the look misplaces by \SI{0.23}{mm}, at a clearance of \SI{0.15}{mm} per side, which corresponds to roughly one pixel of the close view. In physics, the six development failures of our method comprise the three parts with an incorrect feature or receptacle, the part with an internal post, and two simulation failures (\cref{app:failures}).

\subsection{Ablations}
\label{sec:ablation}

\begin{table}[t]
\caption{Ablations (kinematic, strict) on development (Dev.) and held-out (Held.) episodes, with median final error (mm, development). The look uses the height map of the scene cameras or of the close view, for keyed sketches or all parts (ours: close view, all). Bottom: ground-truth sketches.}
\label{tab:ablation}
\centering
\small
\setlength{\tabcolsep}{2pt}
{%
\begin{tabular}{l ccccc c c}
\toprule
Variant & Auto. & Conn. & Board & AIC & Dev. & Held. & Err. \\
\midrule
Sketch only & 32 & 11 & 6 & 1 & 50 & 6 & 0.15 \\
Scene map, keyed & 31 & 13 & 7 & 4 & 55 & 11 & 0.18 \\
Scene map, all & 31 & 13 & 7 & 4 & 55 & 11 & 0.17 \\
Close view, keyed & 33 & 13 & 8 & 4 & 58 & 18 & 0.11 \\
Close view, all & 35 & 13 & 9 & 4 & 61 & 18 & 0.10 \\
\midrule
Oracle, no look & 39 & 13 & 10 & 4 & 66 & 18 & 0.00 \\
Oracle, scene map & 33 & 11 & 10 & 3 & 57 & 18 & 0.10 \\
Oracle, close view & 37 & 13 & 10 & 4 & 64 & 18 & 0.01 \\
\bottomrule
\end{tabular}}
\end{table}

\paragraph{The close view.} We find that the close view extends the benefit of the look from keyed parts to all parts (\cref{tab:ablation}). With a height map from the scene cameras alone, the look corrects the misread rotations of keyed parts but also displaces parts that the sketch had placed precisely, because this map resolves an edge only to approximately \SI{0.4}{mm}. As a result, looking at all parts inserts 55 parts, as many as looking at keyed parts only, and with ground-truth sketches the look loses 9 of 66 insertions. With the close view, the look inserts 58 development parts when it runs for keyed sketches and 61 when it runs for all parts, since it also corrects round parts with openings that the agent fits imprecisely. On the held-out connectors, it inserts all 18 parts instead of 11, and with ground-truth sketches it loses only two parts (\cref{app:failures}).

\subsection{Reusing Sketches}
\label{sec:memoryexp}

We evaluate sketch reuse on eight tasks that cover the characteristic failure modes of the agent (SFP, SC, NEMA 5-15, USB-A, RJ45, D-sub, the D-shaft peg and one AutoMate assembly). For each task, the memory stores the agent runs on the first $k$ of eight placements, and we evaluate it on the last four placements, on which we also run the agent for comparison. Since the memory reuses sketches built by the agent, we expect it to perform as well as the agent whenever the registration is accurate.

\begin{table}[t]
\caption{Reusing sketches on four new placements of eight tasks (32 episodes). The memory stores $k$ agent runs per task and uses the best-registering one or, in the \emph{averaged} variant, their average. Sketch: strict success of the sketch alone. Time: median time to produce the sketch.}
\label{tab:memory}
\centering
\small
\setlength{\tabcolsep}{3pt}
\begin{tabular}{l cccc c}
\toprule
Sketch source & Sketch & Strict & Search & Physics & Time \\
\midrule
Agent & 13 & 30 & 31 & 28 & \SI{90}{s} \\
Memory, $k=1$ & 9 & 30 & 31 & 32 & \SI{3.4}{s} \\
Memory, $k=2$ & 9 & 29 & 30 & -- & \SI{6.6}{s} \\
Memory, $k=4$ & 9 & 28 & 29 & -- & \SI{15.5}{s} \\
\midrule
Averaged, $k=2$ & 4 & 30 & 30 & -- & \SI{3.1}{s} \\
Averaged, $k=4$ & 4 & 29 & 29 & -- & \SI{3.4}{s} \\
\midrule
\textcolor{gray}{Oracle} & \textcolor{gray}{32} & \textcolor{gray}{32} & \textcolor{gray}{32} & \textcolor{gray}{30} & \textcolor{gray}{--} \\
\bottomrule
\end{tabular}
\end{table}

As shown in \cref{tab:memory}, reusing a single stored run inserts 30 of the 32 parts in the strict test, which matches the performance of calling the agent in each episode. The memory produces the sketch in \SI{3.4}{s} instead of about \SI{90}{s} and does not fall back to the agent in any episode. In physics, the memory inserts all 32 parts and the agent 28 (\cref{app:failures}). We find that the look-and-mate step matters even more for reused sketches. A transferred sketch alone inserts only 9 parts, and the look raises this number to 30, because it corrects the rotations and registration offsets that the transferred sketch inherits.

All strict failures of the memory occur on the nearly symmetric SFP and SC connectors. Their sketches mostly arrive flipped, either from a run that read the part flipped or from a yaw error of the registration, and the look then keeps a flipped twin that also fits the opening. Storing more runs does not avoid these failures. Averaged sketches insert only 4 parts alone, likely because the medians mix positions and references from runs with different frames.

\section{Limitations and Conclusion}
\label{sec:limits}

Our evaluation is limited to simulation with exact depth. Real depth sensors can blur millimetre-scale openings even at the distance of the close look, and even exact depth limits fits with a clearance of about one pixel of the close view (\cref{sec:analysis}). Twins with a key below the resolution of the cameras, such as the duplex SC connector, still depend on the rotation that the agent reads.

Our method also assumes a rigid grasp after the look and two extra cameras. The memory helps only on tasks it has seen, and a sketch describes mates reached by a straight motion along one axis, which excludes screw motions and multi-step assemblies.

Our approach suggests that VLM agents for precise manipulation should decide what to measure, leave the measurement and shape matching to geometry, and store their decisions as sketches for reuse.

\section*{Impact Statement}
This paper presents work with the goal of advancing machine learning for robotic manipulation, in particular the assembly of unseen parts. Such systems could lower the cost of automating small-batch assembly and of adapting production lines, which may change the work of people who assemble or set up machines today. Our method runs in simulation, and deploying it on real robots requires the usual safety measures for force-controlled insertion near people.

\bibliography{refs}
\bibliographystyle{plainnat}

\newpage
\appendix
\section{Agent Prompt and Tools}
\label{app:prompt}
\begin{figure}[h]
  \centering
  \includegraphics[width=\textwidth]{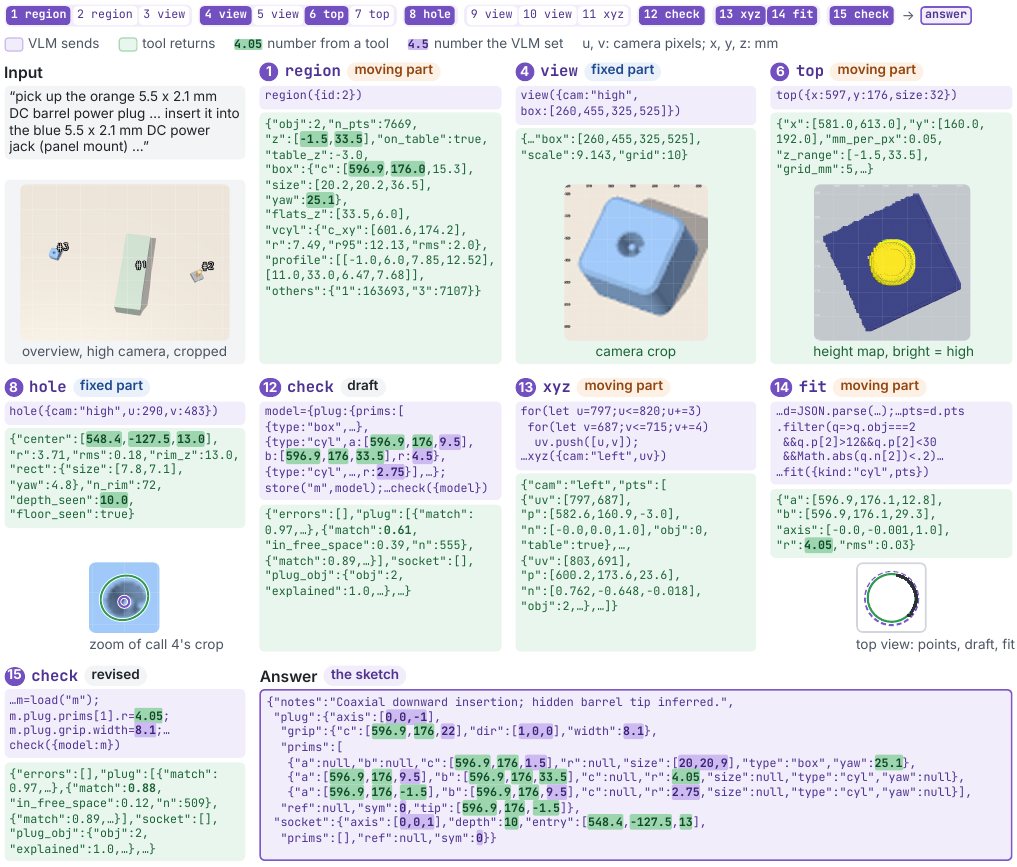}
  \caption{\textbf{The VLM decides what to measure, and most numbers of its sketch come from the tools.} One typical run of GPT-6.1-Sol on a held-out task, a DC barrel plug and its jack. The strip lists the 15 calls of the run, grouped by the code block that sent them, and the cards show the calls with filled chips, with what the VLM sent (purple) and what the tool returned (green), shortened with ``\dots''. Marks show where each number of the sketch comes from, a tool (green) or the VLM (purple). \texttt{check} scores each primitive by the share of its surface seen by the cameras that lies within \SI{2}{mm} of the observed depth (\texttt{match}). The body cylinder of the draft, with a radius set by the VLM itself, scores 0.61, and 0.88 after \texttt{xyz} and \texttt{fit} measure the radius. The insets are drawn from the logs: the pixel sent to \texttt{hole} and the rim it returned on a zoom of the crop of call~4, and the points, draft and fit seen from above.}
  \label{fig:tools}
\end{figure}

\Cref{fig:tools} shows real calls of the tools of \cref{tab:tools} from one run. The agent receives the system prompt below (about 700 tokens, with the instructions, tools and skills of Codex switched off) and a task message. The prompt calls the moving part the plug and the fixed part the socket. The task message contains the instruction, three half-size overview images with object identifiers, the table height and a list of the objects above the table with their centre, footprint and top height. The tools run in a geometric backend that fuses the three depth images into one point cloud, removes the robot with its collision model at the known joint angles, and segments the objects above the table into connected components. The median run makes 17 tool calls. The agent starts with \texttt{region} in 140 of the 141 runs and ends with \texttt{check} in 135, and it calls \texttt{fit} in only a quarter of the runs, likely because \texttt{region} and \texttt{hole} already return fitted shapes. We chose the run of \cref{fig:tools} by a fixed rule. Of the 87 development and held-out runs, 14 call all seven tools and build a sketch that passes the strict test without the look, and we show the one with the median number of calls (15) and the wall time closest to the median of all runs.

{\fontsize{6.5}{7.8}\selectfont
\begin{verbatim}
You build a sketch model of a robot insertion task from RGB-D camera views, for a motion planner. The sketch need not
  look like the parts; it must make the robot grasp the plug and insert it correctly: positions to about 1 mm, axes to
  about 1 degree.

The robot grasps the plug from above and lowers it into or onto the socket without turning it over.

Frame: world, millimetres, z up, robot base at the origin. Pixels: full-resolution camera pixels (the labels of the
  image grids).

Tools (in exec: `r = await tools.mcp__geo__<name>({...})`; print `r.content[0].text`, the JSON result, never `r`
  itself (it holds base64 images); show a returned image with `image(r.content[1])`; batch calls in one exec; no need
  to list the tools):
- view({cam, box?}): image or crop [u0,v0,u1,v1] with a pixel grid
- xyz({cam, uv: [[u,v],...]}): 3D points at pixels (p, normal n, object obj, depth edge)
- region({cam, u, v} or {id}): object at a pixel: z range, upright box, cylinder fit, flat faces, radius profile
- hole({cam, u, v}): opening around a pixel inside it: rim centre, radius, rectangle, rim height
- top({x, y, size?, z0?, z1?}): orthographic height map around world (x, y) mm with a mm grid
- fit({kind: plane|circle|line|cyl|box|pose, pts, model_pts?}): fit 3D points (mm)
- check({model, draw?, cam?, box?}): validate a draft model against the depth data; draw: true adds a drawing

Answer with the model as JSON:
- plug: the part to pick up.
  - prims: cylinders {"type":"cyl","a":[x,y,z],"b":[x,y,z],"r":r} (a, b: end-face centres) and upright boxes
    {"type":"box","c":[x,y,z],"size":[sx,sy,sz],"yaw":deg} that together cover the plug; solid is fine (a ring can be
    one cylinder). Unused fields are null.
  - tip: centre of the plug's end that enters the socket first. axis: unit vector of the plug's motion during
    insertion (out of the tip).
  - sym: rotational symmetry about the axis (0 round, 1 none, n n-fold). ref: unit vector perpendicular to the axis
    that fixes the orientation (e.g. a flat's normal, a key's direction); null when sym is 0.
  - grip: null (the planner grasps across the plug's volume), or {"c","dir","width"} to force a grasp.
- socket: the fixed part.
  - entry: centre of the opening the tip enters (a hole's mouth), or of the top of a post the plug slides over. axis:
    unit vector pointing out of the socket along the insertion line.
  - sym, ref: as for the plug, such that turning plug.ref onto socket.ref (up to the symmetry) mates the parts.
  - depth: how far the tip travels past entry when fully inserted.
  - prims: [] or a coarse cover.
- notes: one short line.

Mating at several features, or a keyed, flat-sided, splined or slotted shape: tip and entry are one corresponding pair
  of points on the main feature, and sym and ref must make every feature line up (read positions and angles in a top
  view). Features hidden under the plug mirror the socket's.

Take numbers from the tools, not from the images. Check the draft with check before answering.

\end{verbatim}
}

\section{Baseline Ports}
\label{app:baselines}
Both baselines were designed for household objects that fill a large part of the image and for grasp-and-place motions with centimetre tolerances. We ported them from their released code, kept all parts that define each method and changed only what our setting requires, as listed below. Both receive the image of the camera with the most top-down view, cropped to the two parts (as their cameras frame the task objects) and upsampled to at least 640 pixels, together with the instance masks of the simulator and the same instruction as our agent. Both use GPT-6.1-Sol with medium reasoning effort through Codex, with code execution switched off, so that each request is a plain VLM call as in the original methods.

\paragraph{ReKep.} We keep the keypoint proposal (DINOv2 ViT-S/14 features inside each mask, PCA to three dimensions concatenated with the normalized 3D position, $k$-means with five clusters per mask, the member closest to each centre, merging by mean shift and numbered boxes on the image), the prompt template, the parser of the constraint program and the subgoal solver (dual annealing with SLSQP over the end-effector pose with the cost terms and weights of ReKep). We make five changes. (i)~We reduce the keypoint merging distance from \SI{6}{cm} to \SI{6}{mm}, because at \SI{6}{cm} a whole connector collapses to a single keypoint. (ii)~We replace the end-effector frame by our gripper frame, which approaches along $z$ instead of along the $x$ axis of Fetch. (iii)~The IK of ReKep counts the descents of the cyclic coordinate descent of Lula, and we emulate it with our damped least-squares solver restarted from random configurations. (iv)~The grasp action advances \SI{1}{cm} along the approach before closing, whereas the Fetch robot of ReKep advances \SI{10}{cm}. (v)~We execute the whole program, so that helper functions that the VLM defines next to the constraints are available. The original parser keeps only the stage functions, which then fail with a name error in most programs of GPT-6.1-Sol. The signed distance field and the collision points come from the meshes of the episode, as in the simulator code of ReKep.

\paragraph{MOKA.} We keep the three prompts (subtask proposal, filtering and motion selection) and the message format, the candidate keypoints (six per object by farthest-point sampling over the mask polygon and its centroid), the $5\times5$ waypoint grid, the parsing of the selected marks (including its indexing), the deprojection with a \SI{1}{cm} offset and the motion. As in the released policy, the gripper moves the function point through the pre-contact waypoint to the target point while pointing down at a fixed yaw. We make three changes. (i)~We replace GroundedSAM by the masks of the simulator. (ii)~We replace the antipodal grasp sampler by a grasp that closes across the narrow side of the moving part at the selected grasp point, which is what an antipodal sampler finds on a standing part. (iii)~We lower the safe height from \SI{25}{cm} to \SI{10}{cm} above the table.

\paragraph{Execution.} For both methods, we take the grasp pose, the pose before the last stage (or \SI{15}{mm} above the final pose if there is none) and the final pose. Our planner plans collision-free paths between these poses. Our compliant controller then moves from the pre-insertion pose through the final pose and up to \SI{30}{mm} beyond it until the part stops, using the same spiral and rotary search at the entry as for our method. The kinematic test removes the error along the insertion axis for all methods, so that a target that ends above the opening is not penalized.

\section{Implementation Details}
\label{app:impl}
\paragraph{Agent.} The VLM runs through the Codex CLI with our system prompt in place of the instructions of Codex, and with the skills, plug-ins, shell and web tools of Codex switched off. A request therefore carries about \num{2700} tokens of fixed overhead in addition to the conversation and the images. The tools run in an MCP server on the geometric backend. We cache each sketch by the content of the snapshot, the prompts, the model and its reasoning effort, so that repeated evaluations reuse it. Over the 141 episodes for which the agent built a sketch, the median run took 17 tool calls, \SI{90}{s}, \num{95}k input tokens (\num{18}k not cached) and \num{2.0}k output tokens.

\paragraph{Planner and controller.} We sample grasps across the primitives of the sketch (12 yaws and several heights, with \SIrange{5}{25}{mm} of the part between the fingers) and try them in three tiers: the grip that the agent suggests, the grasps within 10\% or \SI{1}{mm} of the narrowest width, and the rest. A tier is used only when no grasp of the tiers before it yields a plan. For a round mate, the planner picks the rotation about the axis that is easiest to reach. The fingertips have to stay at least \SI{3}{mm} above the table and, once the part is inserted, \SI{0.5}{mm} away from the points of the fixed part. The robot is modelled by spheres for the arm and convex hulls for the hand and fingers, and the scene by the fused point cloud (\SI{3}{mm} voxels) and the table plane. Paths are planned with RRT-Connect and then shortcut. The controller tracks joint paths by computed torque and inserts with the task-space impedance controller of AutoMate at \SI{500}{N/m} laterally and \SI{150}{N/m} axially. It aligns the part on the planned axis to \SI{0.15}{mm} and descends until the part stops. When the part is blocked at the entry, the controller runs a spiral search (pitch \SI{0.6}{mm}, radius up to \SI{3}{mm}) and a rotary search ($\pm\SI{10}{\degree}$). It stops when the part has bottomed out or travelled \SI{5}{mm} beyond the expected depth.

\paragraph{Look and mate.} The upward camera ($1280\times960$, \SI{60}{\degree} vertical field of view) is mounted in the table. The moving part is presented \SI{12}{cm} above it (about \SI{0.14}{mm} per pixel), with the hand rotated by \SI{90}{\degree} or \SI{180}{\degree} when the rotation of the grasp is out of reach. The wrist camera ($1280\times960$, \SI{45}{\degree} vertical field of view) sits \SI{65}{mm} beside the gripper, across its closing direction, and \SI{85}{mm} behind the fingertips, looking along the approach. In the upward view, we keep the points near the gripper and remove those of the robot with its collision model at the known joint angles, which leaves the moving part. We add the fingers to its profile from their collision geometry. For the close look, the planner moves the wrist camera \SI{18}{cm} above the sketch entry, looking down the axis of the fixed part, at one of eight rotations of the hand about this axis, with the measured part attached for collision checks. Wherever the close view saw a surface or its rays passed through a cell, its height replaces that of the scene cameras. The profile of the moving part and the height map use a \SI{0.25}{mm} grid, and the height map covers at least $\pm\SI{3}{cm}$ around the entry. In the score of \cref{eq:score}, $\tau=\Delta/2$ and $\beta=\SI{1.25}{mm}$. On the GPU, the search evaluates all rotations in \SI{4}{\degree} steps, each turning the part about its tip, and for each rotation all tip positions that keep the centre of the outline within \SI{3}{mm} of its location in the sketch. It then refines the best rotations in \SI{0.5}{\degree} steps and scores each pose at four depths of the tip, from a quarter to all of half the sketched depth and at most \SI{8}{mm}. Fits within 20\% of the best contact count as equivalent. We ignore a sketch rotation that is more than \SI{45}{\degree} away from all fits as a misreading, and the best-scoring fit is selected. When the matched position agrees with the sketch within \SI{0.5}{mm}, we average the two, since the least-squares fits of the agent to rims and walls are more precise than the grid. When the part fits the opening of the sketch only loosely or not at all, we also test the other openings of the fixed part and move the part only if it fits clearly more snugly elsewhere. The robot inserts the part at the matched pose, at least as deep as the sketch planned, along a path planned from the close look. If no path exists, for example because the free rotation of a round part turns the hand into an obstacle beside the fixed part, we search only the position, at the rotation of the plan from the sketch, which the planner has checked for reach and collisions. If no pose is free of penetration, the best score decides. On a desktop GPU, the close look is planned in \SIrange{0.2}{0.3}{s} and the match takes \SIrange{3}{5}{s}, or about \SI{16}{s} when the other openings of a task board are checked.

\paragraph{Choosing the opening by geometry alone.} Without the opening chosen by the VLM, shape matching alone on the map of the scene cameras, which covers the whole board, selects the named receptacle in 7 of the 10 board tasks and inserts 5. It confuses the D-sub, RJ45 and NEMA receptacles, which have narrow internal features that the scene cameras cannot resolve, with looser openings that accept the part. Conversely, it finds the USB-A receptacle that the agent missed. On the board, we therefore keep the opening chosen by the agent unless the part clearly fits better elsewhere.

\paragraph{Search frame and window.} \Cref{tab:window} compares the frame and the size of the search window on the episodes with a keyed sketch, with the height map of the scene cameras alone. When we search over the offset of the gripper, a \SI{3}{mm} window loses five keyed connectors that the gripper holds \SIrange{6.3}{7.5}{mm} away from their mating axis, so that the correct offset lies outside the window. An \SI{8}{mm} window instead lets false fits win, for example on AutoMate part 00083 and on a DisplayPort plug. A window around the tip of the sketch does not depend on the grasp. However, it fails when the agent misreads the rotation of a part with an off-centre tip. For example, the pins of the NEMA 5-15 plug point down into its holder, so the agent infers them from the receptacle and places the sketch tip where it believes the ground pin to be, \SI{6.7}{mm} from the centre. When it misreads the rotation, it also places the entry where this pin would go under its reading. The sketch therefore places the centre of the plug correctly, but the correct pose puts the tip \SIrange{9}{13}{mm} from the entry, outside a tip window, and the best partial overlap keeps an incorrect rotation in all seven misread placements. Bounding the centre of the outline keeps the window small and recovers these rotations. The gripper window still inserts three further placements of the reuse tasks that our window does not, namely two NEMA 5-15 plugs that both windows place about \SI{0.4}{mm} from the assembled pose, on either side of the strict tolerance, and a D-shaft peg with a flat that our match resolves only to \SI{6}{\degree}.

\begin{table}[h]
\caption{Search frame and window of the look (kinematic, strict, height map of the scene cameras) on the episodes with a keyed sketch: development, held-out, and the placements of the reuse tasks that are not in the development set.}
\label{tab:window}
\centering
\small
\begin{tabular}{l ccc}
\toprule
Search & Development (34) & Held-out (14) & Reuse tasks (50) \\
\midrule
Gripper offset, \SI{8}{mm} & 22 & 6 & 35 \\
Gripper offset, \SI{3}{mm} & 18 & -- & -- \\
Tip position, \SI{8}{mm} & 24 & 7 & -- \\
Tip position, \SI{3}{mm} & 25 & 8 & 30 \\
Outline centre, \SI{3}{mm} (ours) & 25 & 8 & 34 \\
\bottomrule
\end{tabular}
\end{table}

\paragraph{Sketch memory.} The object templates of a run are the observed points of the moving and the fixed part from the segmentation of the geometric backend, kept at \SI{1}{mm} spacing in object frames at the centroid of each part on the table. In a new episode, the memory segments the scene with the same backend and considers the objects with a mean colour close to that of a stored part and an extent of at most 1.5 times that of its template plus \SI{1}{cm}. It registers the template of each run to each of these objects by a sweep over the yaw in \SI{15}{\degree} steps, each followed by trimmed ICP in $(x,y,\text{yaw})$ with the height fixed, since both frames lie on the table. The fitness is the share of the points of the scene object within \SI{1.5}{mm} of the template. The run that maximizes the lower of its two part fitnesses provides the sketch, and a fitness below 0.6 makes the memory call the agent. The averaged variant of \cref{tab:memory} registers all runs into the frames of the first run and, among the runs that chose the majority opening, takes the medians of their positions and axes and the majority of their symmetries and references.

\section{Failure Cases}
\label{app:failures}
This section lists the individual failures that \cref{sec:analysis,sec:ablation} summarize.

\paragraph{Baselines.} Even well-formed ReKep constraints relate keypoints that do not lie on the mating features. For example, the VLM defined the axis of a hexagonal peg by two keypoints on its top face, and the solver tilted the peg by \SI{84}{\degree} to align this axis with the vertical direction.

\paragraph{Our method.} The six incorrect rotations of the sketch on development episodes are the NEMA 5-15 plug on the board (\SI{90}{\degree}), the triangular peg (\SI{30}{\degree}), the USB-A and SFP plugs (flipped) and two AutoMate parts (\SI{48}{\degree} and \SI{73}{\degree}), and the look recovers the first four. The three imprecise episodes that fail the strict test even with ground-truth sketches are two loose fits and an asset that is not watertight. The three incorrect locations are an AutoMate part that mates through features on its hidden underside, one that slides along a bar, where the agent took the free end of the bar for the entry, and the USB-A plug in a larger receptacle of the board. In physics, the D-sub plug on the board jams \SI{5.7}{mm} above its seat with all sketches, and a D-shaft peg that starts slightly tilted topples by itself. Ground-truth sketches fail in physics on the part with the post inside and on three parts that jam. On the reuse tasks, the agent loses two SC placements and two D-shaft pegs in physics, which the simulated gripper holds tilted or drops. In the strict test with the close view, they fail on the part with the post and on a loosely mating AutoMate part, which they insert \SI{10}{\degree} from its assembled rotation. On the held-out connectors, the look corrects rotations that the agent misread by \SIrange{77}{106}{\degree} on all three XLR plugs and by \SI{180}{\degree} on two HDMI and two DisplayPort plugs, and five plugs that the sketch offsets by \SIrange{0.2}{0.5}{mm}. For both DisplayPort plugs, a box blocks the grip that the agent suggests, and the planner uses a later grasp tier.

\section{Benchmark Details}
\label{app:tasks}
\begin{figure}[h]
  \centering
  \includegraphics[width=\textwidth]{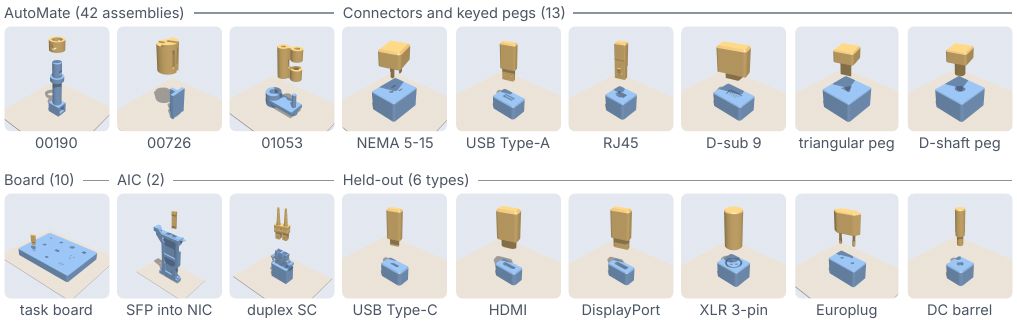}
  \caption{Tasks. Examples of the five task sets, rendered from the assets of the simulator (moving part in orange, lifted above its assembled pose, and fixed part in blue). The held-out connectors were built after the method was fixed and were not used during development.}
  \label{fig:tasks}
\end{figure}
\Cref{fig:tasks} shows examples of the five task sets, and \cref{tab:tasks} lists the procedural and AIC tasks. The procedural parts follow their published dimensions, for example the USB 2.0 drawings, 47~CFR~68.500 for RJ45 and ANSI/NEMA WD~6 for the power plugs. We enlarged the receptacles to clearances of \SIrange{0.12}{0.3}{mm} per side and left out latches and spring contacts. Each part is watertight, does not penetrate its counterpart when seated, and has the declared symmetry along the whole insertion path, which we verified by sweeping rotations at several depths. The tolerances are the largest lateral and rotational errors at which the part still enters, measured on the meshes. The 42 AutoMate assemblies are those in which the moving part stands stably on its mating end. Each episode places both parts at random positions and yaws (the board within $\pm\SI{30}{\degree}$) and adds zero to two box obstacles. The SC adapter and the NIC use the box colliders of the AIC models in the simulator. Because these colliders lack the keys, a flipped SFP module enters its port, which is why the physics test also checks the rotation.

\begin{table}[h]
\caption{Procedural and AIC tasks: rotational symmetry of the mate, clearance per side, lateral and rotational tolerance, and whether the moving part stands on the table or in a holder.}
\label{tab:tasks}
\centering
\small
\begin{tabular}{ll c c c c l}
\toprule
Set & Moving part & Sym. & Clear. (mm) & Lat. (mm) & Rot. (deg) & Start \\
\midrule
Connectors & NEMA 5-15 power plug & 1 & 0.3 & 0.34 & 1.8 & holder \\
 & NEMA 1-15 power plug & 1 & 0.3 & 0.34 & 2.8 & holder \\
 & USB Type-A & 1 & 0.15 & 0.20 & 2.2 & holder \\
 & RJ45 (8P8C) & 1 & 0.15 & 0.20 & 2.1 & holder \\
 & D-sub 9 & 1 & 0.15 & 0.20 & 1.5 & stands \\
 & rectangular peg (Factory) & 2 & 0.124 & 0.17 & 1.3 & stands \\
 & square peg 12 mm & 4 & 0.2 & 0.25 & 2.6 & stands \\
 & hexagonal peg 12 mm & 6 & 0.2 & 0.25 & 4.6 & stands \\
 & triangular peg 14 mm & 3 & 0.2 & 0.25 & 2.3 & stands \\
 & cross peg 14 mm & 4 & 0.2 & 0.25 & 2.1 & stands \\
 & D-shaft peg 12 mm & 1 & 0.2 & 0.25 & 3.7 & stands \\
 & two-pin peg & 1 & 0.2 & 0.25 & 2.0 & holder \\
 & rectangular peg 16$\times$10 mm & 2 & 0.2 & 0.25 & 1.9 & stands \\
\midrule
Board & round peg 8 mm & 0 & 0.2 & 0.25 & 180.0 & stands \\
 & round peg 12 mm & 0 & 0.2 & 0.25 & 180.0 & stands \\
 & round peg 16 mm & 0 & 0.2 & 0.25 & 180.0 & stands \\
 & square peg 12 mm & 4 & 0.2 & 0.25 & 2.6 & stands \\
 & hexagonal peg 12 mm & 6 & 0.2 & 0.25 & 4.6 & stands \\
 & rectangular peg 16$\times$10 mm & 2 & 0.2 & 0.25 & 1.9 & stands \\
 & USB Type-A & 1 & 0.15 & 0.20 & 2.2 & holder \\
 & RJ45 & 1 & 0.15 & 0.20 & 2.1 & holder \\
 & D-sub 9 & 1 & 0.15 & 0.20 & 1.5 & stands \\
 & NEMA 5-15 & 1 & 0.3 & 0.34 & 1.8 & holder \\
\midrule
AIC & SFP module into NIC (2 ports) & 1 & CAD & 0.20 & 1.6 & holder \\
 & duplex SC into adapter & 2 & CAD & 0.44 & 2.6 & holder \\
\midrule
Held-out & USB Type-C & 2 & 0.15 & 0.19 & 3.8 & holder \\
 & HDMI Type A & 1 & 0.15 & 0.20 & 1.7 & holder \\
 & DisplayPort & 1 & 0.15 & 0.20 & 1.5 & holder \\
 & XLR 3-pin & 1 & 0.15 & 0.19 & 2.2 & stands \\
 & Europlug (CEE 7/16) & 2 & 0.25 & 0.30 & 1.7 & holder \\
 & DC barrel 5.5/2.1 mm & 0 & 0.15 & 0.20 & 180.0 & holder \\
\bottomrule
\end{tabular}
\end{table}

\end{document}